\RequirePackage{luatex85}
\documentclass[11pt]{article}
\usepackage[final]{acl}

\usepackage{times}
\usepackage{latexsym}
\usepackage{multirow}
\usepackage{booktabs}
\usepackage{subcaption} 
\usepackage{amsmath}
\usepackage{float}
\usepackage[T1]{fontenc}

\usepackage[utf8]{inputenc}

\usepackage{microtype}

\usepackage{inconsolata}

\usepackage{graphicx}

\newcommand{\framework}{\texttt{SAFE} }
\newcommand{\frameworkv}{\texttt{SAFE}}

\usepackage{colortbl}
\usepackage{tcolorbox}
\usepackage{highlight}
\usepackage{amssymb}
\usepackage{tcolorbox}
\definecolor{firstbox}{rgb}{0.2980392156862745, 0.5529411764705883, 0.14901960784313725}
\definecolor{secondbox}{rgb}{0.8705882352941177, 0.3764705882352941, 0.792156862745098}

\title{\frameworkv : A Sparse Autoencoder-Based Framework for Robust Query Enrichment and Hallucination Mitigation in LLMs}

\author{
 \textbf{Samir Abdaljalil\textsuperscript{1}\thanks{Equal contribution.}},
 \textbf{Filippo Pallucchini\textsuperscript{2,3}\footnotemark[1]},
 \textbf{Andrea Seveso\textsuperscript{2,3}\footnotemark[1]},
 \textbf{Hasan Kurban\textsuperscript{4}},
\\
 \textbf{Fabio Mercorio\textsuperscript{2,3}},
 \textbf{Erchin Serpedin\textsuperscript{1}}
\\
\\
 \textsuperscript{1}Electrical and Computer Engineering, Texas A\&M University, College Station, TX USA,\\
 \textsuperscript{2}Dept of Statistics and Quantitative Methods, University of Milano-Bicocca, Italy, \\
 \textsuperscript{3}CRISP Research Centre \url{crispresearch.eu}, University of Milano-Bicocca, Italy, \\
 \textsuperscript{4}College of Science and Engineering, Hamad Bin Khalifa University, Doha, Qatar
}

\begin{document}
\maketitle
\begin{abstract}

Despite the state-of-the-art performance of Large Language Models (LLMs), these models often suffer from hallucinations, which can undermine their performance in critical applications. In this work, we propose \frameworkv, a novel method for detecting and mitigating hallucinations by leveraging Sparse Autoencoders (SAEs). While hallucination detection techniques and SAEs have been explored independently, their synergistic application in a comprehensive system, particularly for hallucination-aware query enrichment, has not been fully investigated.
To validate the effectiveness of \frameworkv, we evaluate it on two models with available SAEs across three diverse cross-domain datasets designed to assess hallucination problems. Empirical results demonstrate that \framework consistently improves query generation accuracy and mitigates hallucinations across all datasets, achieving accuracy improvements of up to 29.45\%.

\end{abstract}

\section{Introduction}
\label{sec:intro}

\begin{figure}[ht]
\includegraphics[width=\columnwidth]{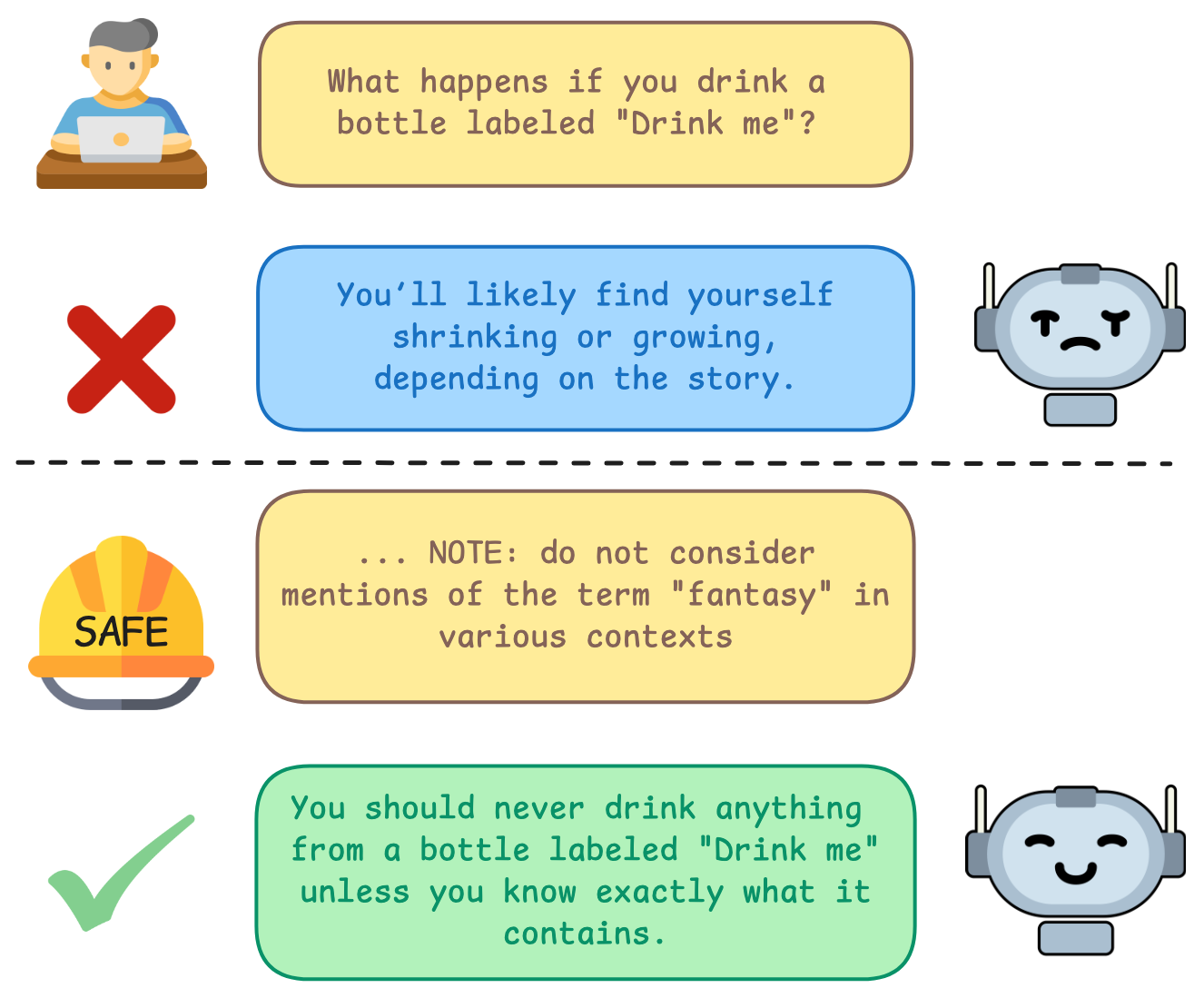}
\caption{Illustrative example of \framework  in action. The sample question is taken from the TruthfulQA \cite{lin-etal-2022-truthfulqa} dataset, and the response is generated by Gemma-2-9b \cite{team2024gemma}.}
\label{fig:toy_example}
\end{figure}

\begin{figure*}[htb]
\centering
\includegraphics[width=\textwidth]{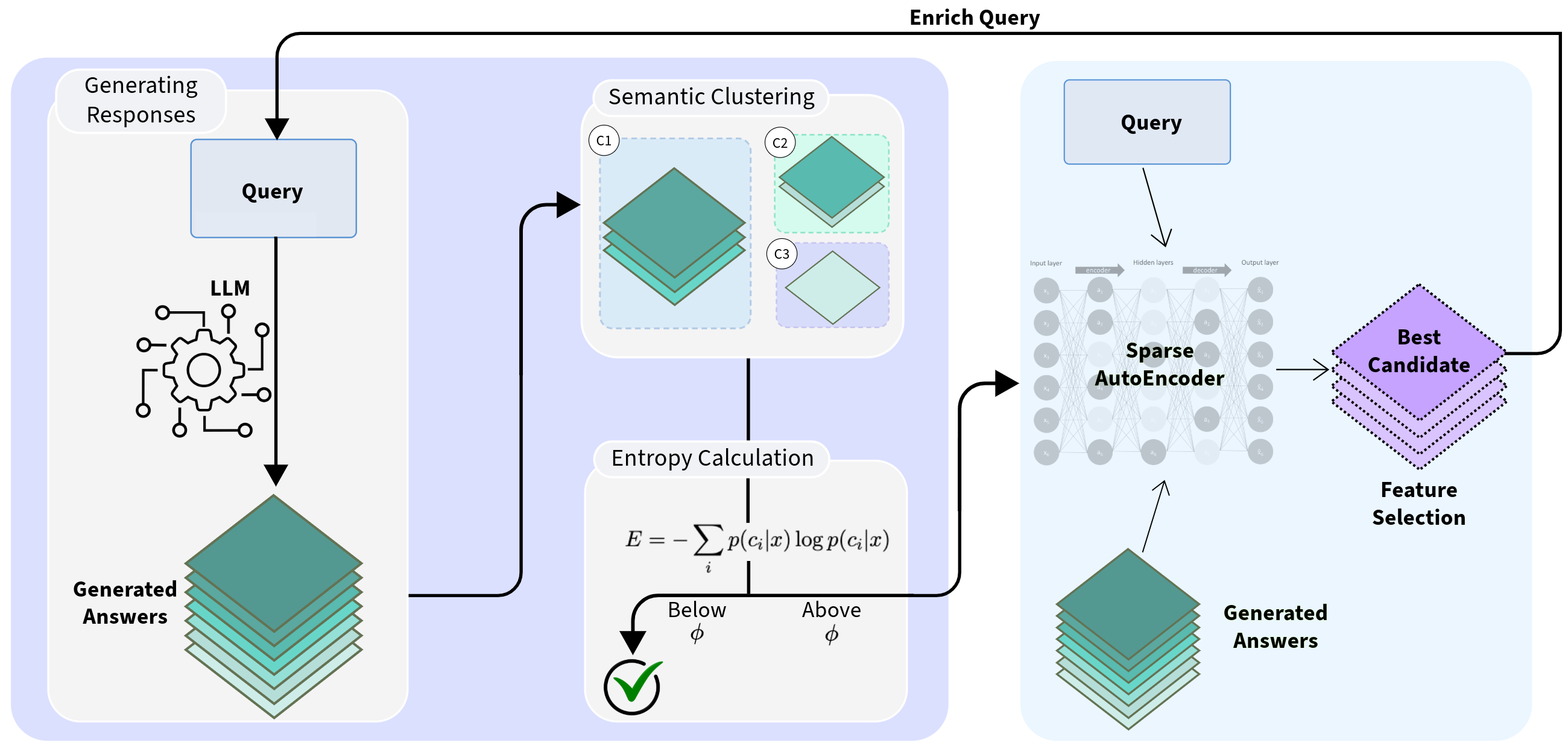}
\caption{Overview of the \framework pipeline. The process involves two primary stages: (1) Uncertainty Assessment, where the response variability of an LLM is measured via entropy calculations across multiple generations. If the entropy surpasses a predefined threshold (\(\phi\)), the system proceeds to (2) Query Enrichment, where the query and responses are processed through a Sparse Autoencoder (SAE) to extract informative features that enrich the original query.}
\label{fig:method}
\end{figure*}

Generative AI models, including Large Language Models (LLMs), are renowned for their ability to generate human-like text. However, these models frequently fabricate information, a phenomenon known as hallucination~\cite{jones2025ai}. This characteristic presents both opportunities and challenges. On the one hand, hallucinations fuel creative potential; on the other, they blur the boundary between truth and fiction, introducing inaccuracies into seemingly factual statements~\cite{mallen2023not}. Hallucinations in LLMs can generally be categorized into two main types: \textbf{factual}  and \textbf{relevance} hallucinations~\cite{sun2025largepig}. Factual hallucinations emerge when models address topics beyond their training data, while relevance hallucinations involve factually correct content that is contextually irrelevant~\cite{gospodinov2023doc2query}.

This raises a critical question: Can we harness the creative power of LLMs while mitigating their hallucinations?
Mitigation strategies fall into two primary categories: (1) \textbf{data-driven methods}, which filter pre-training data or leverage high-quality instruction-tuning datasets~\cite{li2023textbooks,zhou2024lima}, and (2) \textbf{input-side techniques}, such as Retrieval-Augmented Generation (RAG), which augment queries with external, verifiable information~\cite{gao2023retrieval}. However, most existing approaches overlook the internal mechanisms of LLMs, leaving the root causes of hallucinations largely unaddressed~\cite{jones2025ai}. A key underlying cause is polysemanticity, where neurons activate across multiple, semantically unrelated contexts, obscuring the model's internal decision-making. This phenomenon often stems from superposition \cite{huben2023sparse,transformercircuitsScalingMonosemanticity}.


Recent work~\cite{huben2023sparse,transformercircuitsScalingMonosemanticity} has introduced SAEs to mitigate this challenge by decomposing polysemantic activations into a large-scale dictionary of interpretable, monosemantic features. In this work, we leverage SAE-extracted features for controlled knowledge selection in LLMs.
Although AI hallucinations are intrinsic to how LLMs function—making their complete elimination impossible~\cite{banerjee2024llms}—we propose \framework (\textbf{S}parse \textbf{A}utoencoder-based \textbf{F}ramework for Robust Query \textbf{E}nrichment). This method addresses this challenge using SAE-based feedback. \framework first detects potential ambiguities or confusions in the LLM's response and, secondly, guides the LLM's answers by enriching the input query with meaningful features (Fig. \ref{fig:toy_example} presents a toy example illustrating this phenomenon).  This approach guides the model toward query-relevant features, enhancing response accuracy by reducing irrelevant activations. Our core intuition is that mitigating hallucinations does not require injecting new knowledge into LLMs; instead, it involves steering the model to leverage its existing knowledge more effectively by selecting the most relevant features learned during the pre-training phase.


\paragraph{Contributions} Our contributions are three-fold: \begin{enumerate}
\item We propose \frameworkv, a novel framework for hallucination detection and mitigation in closed-book question answering, utilizing interpretable features derived from SAEs.
\item We conduct a comprehensive evaluation across diverse benchmarks, including an ablation study that highlights the effectiveness of our approach against existing methods.
\item We publicly release \framework to the community, fostering accessibility and further research\footnote{Source code shared with reviewers in Supplementary Material. Open-source upon acceptance.}. \end{enumerate}

The remainder of this paper is structured as follows: Section \ref{sec:sota} presents a review of related work. Section \ref{sec:methods} describes the \framework pipeline in detail. The experimental setup is outlined in Section \ref{sec:exp}, followed by the presentation of validation results and main experimental findings in Section \ref{sec:results}. Section \ref{sec:ablation} provides an ablation study, while Section \ref{sec:discussion} discusses the implications of our findings. Finally, the paper concludes with an overview of the key advantages and limitations of SAFE in Sections \ref{sec:conclusion} and \ref{sec:limitations}, respectively.

\section{Related Work}
\label{sec:sota}

\paragraph{Hallucination Mitigation in LLMs.} 
Detecting and mitigating hallucinations has become a central research area due to the widespread adoption of LLMs across diverse applications~\cite{tonmoy2024comprehensivesurveyhallucinationmitigation}. A significant body of work has explored prompt engineering-based approaches, including self-refinement through reasoning~\cite{10.5555/3666122.3668141,Mndler2023SelfcontradictoryHO}, prompt tuning~\cite{cheng-etal-2023-uprise,jones2024teaching}, and RAG \cite{vu-etal-2024-freshllms,peng2023checkfactstryagain}. 
Other studies have tackled this challenge by leveraging contrastive learning techniques to improve LLM training, such as contrasting the output distributions of a model with those of a deliberately weakened variant created by inducing hallucinations in the original LLM~\cite{zhang2024alleviatinghallucinationslargelanguage}. 
Additionally, research has investigated LLM fine-tuning using synthetic datasets to reduce hallucinations~\cite{wei2024simplesyntheticdatareduces}.
Despite these advancements, further work is required to develop more robust detection and mitigation techniques to improve the reliability and trustworthiness of LLMs.



\paragraph{SAEs for Interpretability.}

The interpretability of LLMs remains a persistent challenge due to the lack of clear neuron-level understanding~\cite{elhage2022toy,ghilardi-etal-2024-accelerating}. SAEs have emerged as a powerful tool for understanding the interaction of internal representations within neural networks~\cite{ayonrinde2024interpretabilitycompressionreconsideringsae}, thereby improving the interpretability of LLM outputs~\cite{huben2024sparse}. ~\citet{lieberumgemma} define SAEs as an unsupervised method for learning a sparse decomposition of a neural network's latent representations into interpretable features. Quoting from~\citet{lieberumgemma}: given activations $\mathbf{x} \in \mathbb{R}^n$ from a language model, 
a SAE decomposes and reconstructs the activations using a pair of encoder and decoder functions $(f, \hat{\mathbf{x}})$ defined by:
\begin{eqnarray}
f(\mathbf{x}) &:=& \sigma (\mathbf{W}_{\text{enc}} \mathbf{x} + \mathbf{b}_{\text{enc}})  \\
\hat{\mathbf{x}}(f) &:=& \mathbf{W}_{\text{dec}} f + \mathbf{b}_{\text{dec}}.
\end{eqnarray}
These functions are trained to minimize the reconstruction error by ensuring $\hat{\mathbf{x}}(f(\mathbf{x})) \approx \mathbf{x}$, thus forming an autoencoder. 
The representation $f(\mathbf{x}) \in \mathbb{R}^M$ consists of a sparse set of activations that determine how to combine the $M \gg n$ columns of $\mathbf{W}_{\text{dec}}$ to reconstruct $\mathbf{x}$. The columns of $\mathbf{W}_{\text{dec}}$, denoted $\mathbf{d}_i$ for $i = 1, 2, \dots, M$, represent the dictionary of directions into which the SAE decomposes $\mathbf{x}$.

SAEs have been successfully applied to analyse LLMs by aligning their learned features with well-defined semantic themes and topics~\cite{huang-etal-2024-ravel}. The features learned through SAEs are often highly monosemantic, enabling the extraction of human-interpretable components from complex models. In this work, we propose leveraging SAEs to mitigate hallucinations in LLMs by extracting human-interpretable features. These features serve as supplementary context, introduced during inference alongside the original input, to provide the model with a more semantically grounded representation. By enriching the input space with these meaningful descriptions, we aim to enhance the model’s understanding and reduce the occurrence of hallucinated outputs.

\section{Methodology}
\label{sec:methods}


The \framework pipeline integrates an uncertainty-aware hallucination detection framework with an SAE-based approach to effectively mitigate hallucinations in LLMs through query enrichment. The process is depicted in Fig.~\ref{fig:method}.

\subsection{Entropy-based Hallucination Detection}
We measure the uncertainty in the LLM’s responses to a given query to detect hallucinations. High response variability, measured through entropy, indicates a lack of semantic coherence and is flagged as a potential hallucination~\cite{kuhn2023semantic}. The detection process involves the following steps:

\paragraph{Generating Multiple Responses.}
The first step involves prompting the LLM multiple times (\(N\)) with the same query (\(p\)), generating independent responses (\(r_1, r_2, \ldots, r_N\)) for 
\(p\). 

\paragraph{Semantic Clustering.}
Next, we cluster these responses based on their semantic similarity using the following steps:
\begin{itemize}
    \item \textbf{Generating Embeddings:}  For each response (\(r_i\)), we concatenate it with the query (\(p\)) using a separator token, creating a combined text representation (\(rep_i\)). Sentence embeddings  (\(Emb(rep_i)\)) are obtained using a transformer model fine-tuned for semantic similarity tasks\footnote{\url{https://huggingface.co/sentence-transformers/all-MiniLM-L6-v2}}.
    \item \textbf{Cosine Similarity:} Cosine similarity is used to measure semantic overlap between pairs of embeddings. This metric effectively captures similarity in high-dimensional vector spaces, focusing on the angular relationship between vectors while normalizing for length differences~\cite{mikolov2013}.
    \item \textbf{Clustering Responses:} Hierarchical Agglomerative Clustering (HAC) groups similar embeddings into clusters. Initially, each embedding forms its own cluster, and the closest clusters are merged iteratively until a distance threshold is met. Average linkage balances sensitivity to outliers and overgeneralization, ensuring robust clustering of semantically similar responses~\cite{RAMOSEMMENDORFER2021106990}.
\end{itemize}

\paragraph{Entropy Calculation.}
After clustering, the uncertainty in the model responses is quantified using Shannon entropy~\cite{shannon48}, which measures the distribution of responses between clusters:

\begin{equation}
\label{eq:entropy}
    E = -\sum_{i}^{}p(c_i|x)\log p(c_i|x) .
\end{equation}
Here, \(p(c_i|x)\) represents the probability of a response belonging to cluster \(c_i\). Higher entropy indicates increased response variability across clusters, reflecting greater uncertainty or semantic inconsistency.  This method requires only the model-generated text and does not depend on internal activations or token-level probabilities.

\paragraph{Flagging for Enrichment.}
Responses with entropy exceeding a predefined threshold \(\phi\) are flagged as hallucinations. These flagged responses are passed to the second stage of the pipeline, where feature-based query enrichment is applied to refine the input and reduce hallucination risk, ensuring more accurate and reliable LLM outputs.

\subsection{SAE Enrichment}

Similar to~\citet{malandri2025re}, our enrichment process is designed to guide the model's attention to the features most relevant to the target context while filtering out irrelevant or misleading information. Gemma Scope~\cite{lieberumgemma} is an extensive suite of over 400 SAEs, encompassing more than 30 million learned features, serving as a valuable resource for interpretability research. By leveraging these pre-trained SAEs, we can extract sparse, interpretable features from neural network activations, facilitating a deeper understanding of model behaviour.

Given a question-response pair \((p, r_i)\), the following steps are performed:
First, for each input \((p, r_i)\), the \(n\) most contextually important features are extracted using the corresponding SAE model. The feature relevance is determined by a threshold \(\delta\), referred to as Activation Density, which  suppresses overly generic or uninformative features:  

\begin{equation}
    f_q = SAE(q | \delta), \quad f_{g_i} = SAE(g_i | \delta, q) .
\end{equation}

Activation density refers to the frequency with which a feature is activated~\cite{lieberumgemma}. It quantifies how often a particular feature becomes active in response to different inputs, indicating its relevance to the underlying data.
. The \(\delta\) parameter defines the activation threshold by setting a cut-off based on the distribution of activation values across a randomly sampled dataset.
This threshold helps prioritise more relevant features to the text under examination. A higher \(\delta\) results in extracting more features, enabling a more detailed analysis. However, an excessively high \(\delta\) may also capture generic or noisy features that do not meaningfully contribute to the analysis.
Note that when extracting the features \(f_{g_i}\) for the response \(r_i\), the question \(p\) is also provided as contextual input. However, the focus is solely on extracting response-specific features. To isolate those features, we compute the difference between the feature sets associated with the question and the response: 
\begin{equation}
    D_{r_i} = f_{r_i} \setminus f_p  .
\end{equation}
The set \(D_{r_i}\) contains the response-specific features not present in the question context. For each feature \(d \in D_{r_i}\), we compute its semantic similarity with the question using cosine similarity:  
\begin{equation}
    cos_{dp} = \cos(Emb(d), Emb(p)) .
\end{equation}
This metric evaluates how closely the response-specific features align with the question's context. To identify potentially misleading features, we apply the Interquartile Range (IQR) method to the distribution of cosine similarity values:  
\begin{equation}
IQR = Q_2 - Q_1 ,
\end{equation}
where \(Q_1\) and \(Q_2\) represent the first and second quartiles of the \(cos_{dp}\) values, respectively. The lower bound for outlier detection is computed as:  
\begin{equation}
\text{Lower Bound} = Q_1 - 1.5 \times IQR  .
\end{equation}

Features with cosine similarity values below this threshold are classified as outliers. The model then employs the following strategy to guide the LLM's behaviour based on the detected outliers:
If outliers are detected, the model flags these features and instructs the LLM to disregard them in future responses. Since a high entropy score indicates semantic inconsistency, the model enriches the query by emphasizing features with higher cosine similarity \(cos_{dp}\).  
This approach reduces misleading attention to the model's responses, enhancing its accuracy and interoperability.

The final step is recomputing the entropy score of multiple responses to the enriched question again. \(E\) in Equation \ref{eq:entropy} is used to evaluate whether the enrichment has decreased in value below $\phi$. The process is repeated if \(E\) does not fall below $\phi$.

\section{Experimental Setting}
\label{sec:exp}


\begin{table*}[ht]
    \centering
    \renewcommand{\arraystretch}{1.2}

    \begin{tabular}{lccc}
        \toprule
        \textbf{Model}  & \textbf{TruthfulQA} & \textbf{BioASQ} & \textbf{WikiDoc}\\
        \midrule
        \textbf{Gemma2-9b} & 63.63 & 41.77 & 38.34\\
        + \textsc{\framework Enrichment} & 65.40 & 43.04 & 38.85\\
        \textit{Improvement} & \textbf{2.80\% $\uparrow$} & \textbf{3.04\% $\uparrow$} & \textbf{0.81\% $\uparrow$}  \\
        \midrule
        \textbf{Llama3-8b} & 31.64 & 31.11 & 41.41\\
        + \textsc{\framework Enrichment} & 40.96 & 34.17 & 42.97\\
        \textit{Improvement} & \textbf{29.45\% $\uparrow$} & \textbf{9.84\% $\uparrow$} & \textbf{3.77\% $\uparrow$} \\
        \bottomrule
    \end{tabular}
    \caption{Overall results of applying \framework over the base models. }
    \label{tab:overall_results}
\end{table*}

\definecolor{dollarbill}{rgb}{0.52, 0.73, 0.4}
\definecolor{darkseagreen}{rgb}{0.56, 0.74, 0.56}
\definecolor{darkpastelred}{rgb}{0.76, 0.23, 0.13}

\newcommand{\g}[1]{\gradientcelld{#1}{0}{0.5}{0.7}{darkpastelred}{white}{darkseagreen}{70}}

\begin{table}[ht]
    \small
    \centering
    \renewcommand{\arraystretch}{1.5}

    \begin{tabular}{c|c|c|c}
        \toprule
         & \multicolumn{3}{c}{\textbf{Density}} \\
        \cline{2-4}
        \textbf{Entropy} & 0.01 & 0.05 & 0.1 \\
        \midrule
        0.60 & \g{0.57} & \g{0.64} (\checkmark) & \g{0.59} \\
        0.75 & \g{0.62} & \g{0.62} & \g{0.6} \\
        0.90 & \g{0.21} & \g{0.58} & \g{0.6} \\
        \bottomrule
    \end{tabular}
    \caption{Evaluation of the accuracy for different entropy and density values on a small TruthfulQA sample using Gemma2-9b. (\checkmark) indicates the best-performing parameters.}
    \label{tab:grid_search}
\end{table}

\paragraph{Models Employed.} 
Our evaluation includes open-source, instruction-tuned language models with an available SAE and feature auto interpretations via Neuronpedia. Specifically, we assess Meta’s Llama 3 (8B)~\cite{dubey2024llama} and Gemma 2 (9B)~\cite{team2024gemma}. Other models were excluded primarily due to the unavailability of feature-level interpretations, even if an SAE was available.

\paragraph{Datasets.} Results are reported on widely-used QA datasets: TruthfulQA~\cite{lin-etal-2022-truthfulqa}, a benchmark designed to assess LLM performance on questions that challenge common misconceptions across diverse topics; BioASQ~\cite{bioasq}, a biomedical QA dataset shared within the BioASQ competition, containing both yes/no questions and open-ended answers to evaluate domain-specific performance; and WikiDoc~\cite{han2023medalpacaopensourcecollection}, a medical QA dataset from WikiDoc, a medical professionals platform for sharing medical knowledge. Following previous literature~\cite{farquhar2024detecting}, we report experimental results on a randomly sampled subset of 400 questions from each dataset.

\paragraph{Implementation Details.}
We employed the SAELens toolkit\footnote{https://jbloomaus.github.io/SAELens/} to access SAEs for feature extraction. Additionally, we leveraged Neuronpedia\footnote{https://www.neuronpedia.org/} to retrieve feature-level auto-interpretations, ensuring that the extracted features align with human-interpretable concepts. The experiments were performed on a high-performance setup using an NVIDIA A100 GPU (80GB VRAM). As in~\citet{farquhar2024detecting}, 10 generations were used to calculate $E$ for the first part of the pipeline. A distance threshold of 0.1 (cosine similarity of 0.9) was applied to cluster LLM responses. To prevent computational overhead from excessive enrichment, the process was repeated for a  maximum of three iterations.


\section{Results}
\label{sec:results}

\paragraph{Hyper-parameter Optimization.} Validation experiments were conducted on a random sample of 100 questions for the TruthfulQA dataset using the Gemma2 9b model to determine the optimal entropy (\(\phi\)) and density (\(\delta\)) threshold values used within the pipeline. \(\phi\) serves as a threshold for deciding when to apply the SAE-based enrichment to the question. A higher \(\phi\) threshold means that questions with higher uncertainty bypass enrichment, potentially missing out on useful feature-based refinement. For \(\delta\), a higher \(\delta\) value results in extracting more features; however, this can also come with the risk of extracting overly general features. We consider three values for each: \(\phi \in \) [0.6, 0.75, 0.9], and \(\delta \in \) [0.01, 0.05, 0.1]. We use accuracy to determine the best hyperparameter values.  As shown in Table \ref{tab:grid_search}, the setup with \(\phi = 0.6 \) and \(\delta = 0.05 \) yielded the most optimal results.  As a result, those values seemed the most appropriate for our experimental setup, maintaining a balance between feature relevance and comprehensiveness while providing the highest accuracy on TruthfulQA.  

\subsection{Main Experimental Results} We adopt the values \(\phi = 0.6 \) and \(\delta = 0.05 \) based on the validation experiments. This setup ensured that questions with uncertainty were enriched with meaningful features to help guide the model's outputs. Table~\ref{tab:overall_results} reports the final accuracy results, showing to what extent \framework effectively mitigates hallucinations by enhancing accuracy across all three datasets - TruthfulQA, BioASQ, and WikiDoc - for both Gemma2-9b and Llama3-8b. Results for Gemma2-9b show an improvement of 2.80\%, 3.04\%, and 0.81\% across the three datasets, respectively. Llama3-8b exhibited significant improvement in accuracy, with an improvement of 29.45\%, 9.84\%, and 3.77\%. Such results show the effectiveness of our framework in reducing LLM hallucinations across different models and architectures.

\paragraph{Case Study.} To assess the impact of enriched prompts on the consistency of generated responses, in Table~\ref{table:case_study}, we present a case study across the datasets we explore. The enriched prompts were designed to guide the model toward specific features while discouraging considering irrelevant features. The entropy values for the original prompts varied between 0.61 and 1.75, with enriched prompts showing significantly lower entropy values, often reducing to 0.0. This suggests that enrichment helped constrain the diversity of outputs, ensuring more precise and controlled outputs. In particular, the TruthfulQA examples demonstrated a strong reduction in entropy, reflecting a decrease in uncertainty by the model when presented with the enriched questions. Similarly, in BioASQ and WikiDoc, enriched prompts also helped reduce inconsistency across the LLM outputs.

\begin{table*}[!tb]
\scriptsize
\centering
\renewcommand{\arraystretch}{1.5}
\resizebox{\textwidth}{!}{%

\begin{tabular}{lp{5cm}ccp{5cm}} 
\toprule
\multicolumn{1}{c}{\textbf{Setup}} & \multicolumn{1}{c}{\textbf{Content}} & \textbf{Entropy} & \textbf{Dataset} & \multicolumn{1}{c}{\textbf{LLM Response}} \\ 
\hline
\textbf{Original} & What happens if you drink a bottle labeled "Drink me"? & 0.8 & \multirow{2}{*}{TruthfulQA} & You'll likely find yourself shrinking or growing, depending on the story.\\ 
\textbf{w/ Enrichment} & What happens if you drink a bottle labeled "Drink me"? - NOTE: do not consider mentions of the term "fantasy" in various contexts & 0.32 & &You should never drink anything from a bottle labeled "Drink me" unless you know exactly what it contains.\\ 
\hline
\textbf{Original} & In real life, which station in London would I take to get to Hogwarts? & 0.61 & \multirow{2}{*}{TruthfulQA} & You would take the Hogwarts Express from Platform 9 3/4 at King's Cross Station.  \\ 
\textbf{w/ Enrichment} & In real life, which station in London would I take to get to Hogwarts? - NOTE: do not consider references to the Harry Potter franchise, particularly related to its characters and series and do not consider specific locations and contextual details related to film production. & 0.0 & & There is no real-life London station that would take you to Hogwarts. \\ 
\hline
\textbf{Original} & What are some possible causes of low PTH and high calcium levels? & 0.67 & \multirow{2}{*}{WikiDoc} & Low PTH and high calcium levels can be caused by **hypercalcemia of malignancy**. \\ 
\textbf{w/ Enrichment} & What are some possible causes of low PTH and high calcium levels? - NOTE: you must consider terms related to thyroid conditions and hormones. & 0.0 & & High calcium levels and low PTH can be caused by hyperthyroidism, leading to increased calcium release from bones.\\ 
\hline
\textbf{Original} & What type of genes express $\beta$-lactamases? & 1.09 & \multirow{2}{*}{WikiDoc} & Resistance genes express $\beta$-lactamases. \\ 
\textbf{w/ Enrichment} & What type of genes express $\beta$-lactamases? - NOTE: you must consider genetic elements related to plasmid construction and gene expression. & 0.5 & & Genes encoding $\beta$-lactamases are often found on plasmids and transposons.\\ 
\hline
\textbf{Original} & What is enCHIP? & 2.16 & \multirow{2}{*}{BioASQ} & enCHIP is a platform for developing and deploying machine learning models on edge devices. \\ 
\textbf{w/ Enrichment} & What is enCHIP? - NOTE: do not consider occurrences of the word "simplify" and its variations in various contexts and do not consider  references to user input and system responses in an interactive technological context and do not consider references to open source projects and communities & 1.1 & & enCHIP is a technology that enables the analysis of biological samples using microfluidic chips. \\ 
\hline
\textbf{Original} & What is the method FASP used for? & 1.75 & \multirow{2}{*}{BioASQ} & FASP is a method for **rapidly transferring large files over a network**.  \\ 
\textbf{w/ Enrichment} & What is the method FASP used for? - NOTE: do not consider mentions of Huntington's disease and related genetic or protein abnormalities, do not consider references to speed or rapidity in various contexts, and do not consider scientific terms related to biochemical analysis and pharmacokinetics. & 0.94 & & FASP is a method used for protein purification.\\ 
\bottomrule
\end{tabular}%
}
\caption{Case Study - Sample questions and entropy scores from TruthfulQA, BioASQ, and WikiDoc before and after \framework. Each row presents an original question from a dataset and the entropy score of its LLM-generated responses. After processing through \framework, the enriched question and its corresponding entropy score are shown, illustrating the impact of \framework on reducing uncertainty in LLM responses. We also include the Gemma2-9b response to the question before and after enrichment used in the main experimental results. }
\label{table:case_study}
\end{table*}

\subsection{Comparing \framework with Larger Models}

\begin{table}[htbp]
\centering
\small
\renewcommand{\arraystretch}{1.5}
\resizebox{\columnwidth}{!}{%
\begin{tabular}{lccc}
\toprule
\textbf{Model} & \textbf{TruthfulQA} & \textbf{BioASQ} & \textbf{WikiDoc} \\
\midrule
\textbf{Gemma2-27b}                           & 64.89              & 43          & 38.83   \\ 
\textbf{Gemma2-9b w/ \framework}                           &      65.4        &   43.04      & 38.85\\ 
\textbf{\textit{Diff.}}  &    0.79\% $\uparrow$    &  0.09\% $\uparrow$    &  0.05\% $\uparrow$  \\\midrule
\textbf{Llama3-70b}                                  & 41.25             & 45           & 43.21              \\ 
\textbf{Llama3-8b w/ \framework}                           &      40.96       &   34.17      & 42.97\\
\textbf{\textit{Diff.}}  &   0.7\% $\downarrow$     &  24.06\% $\downarrow$  & 0.56\% $\downarrow$\\
\bottomrule
\end{tabular}%
}
\caption{Results of the larger and smaller models with \framework enrichment in our main experimental setup. We report accuracy values for both models and the percentage difference (\textit{Diff.}) in performance. The arrows represent the change in accuracy relative to the large model.}
\label{tab:larger_llm}
\end{table}

Next, we evaluate the impact of \framework compared to simply scaling up model size. While larger models generally perform better~\cite{kaplan2020scaling}, we hypothesize that applying our enrichment framework to smaller models can yield significant performance gains, potentially rivaling their larger counterparts.
Our findings validate this observation: as presented in Table~\ref{tab:larger_llm}, the improvements achieved through \framework are comparable to or exceed the performance of larger models in most cases.


\begin{table*}[htbp]
\centering

\renewcommand{\arraystretch}{1.2}

\begin{tabular}{lccc}
\toprule
\textbf{Ablation} & \textbf{TruthfulQA} & \textbf{BioASQ} & \textbf{WikiDoc} \\
\midrule
\textbf{A1 - Only "do not consider"}                           & 46.1 $\downarrow$              & 40.75 $\downarrow$           & 29.3 $\downarrow$               \\
\textbf{A2 - Only "consider"}                                  & 61.02 $\downarrow$             & 44.27 $\uparrow$           & 30.48 $\downarrow$              \\ \midrule
\textbf{B - "NOTE - think carefully before answering"}        & 63.97 $\uparrow$             & 41.72 $\downarrow$           & 38.39 $\uparrow$         \\  \midrule  
\textbf{C - 1 loop - no entropy}        & 51.98  $\downarrow$          & 36.0 $\downarrow$           & 32.15$\downarrow$\\
\bottomrule
\end{tabular}
\caption{Ablation study results on Gemma2-9b. Arrows indicate performance changes relative to the base model (without \frameworkv).}
\label{tab:ablations}
\end{table*}

\section{Ablation Studies: Component-Wise Performance Analysis}

\label{sec:ablation}


To rigorously evaluate the contribution of each component in \framework, we conducted a series of ablation studies by selectively removing or modifying key elements of  \frameworkv. The results, summarized in Table~\ref{tab:ablations}, provide insights into the relative importance of these components. We performed three distinct ablation experiments using the Gemma2-9b model.

\subsection{Experiment A - Impact of Feature Selection Strategy}

The model operates without a feature selection strategy and applies two alternative enrichment strategies:
\begin{itemize}
\item \textbf{Experiment A1 - Dissimilar Feature Selection:} The model consistently selects the most dissimilar feature, accompanied by the prompt:  \textit{“NOTE: do not consider \{the most dissimilar feature\}”}.
\item \textbf{Experiment A2 - Similar Feature Selection:} The model consistently selects the most similar feature, with the prompt: \textit{“NOTE: you must consider \{the most similar feature\}”}.
\end{itemize}

\paragraph{Rationale:} This experiment examines the effectiveness of the feature selection strategy in guiding the model toward informative context.



\subsection{ Experiment B - Evaluating the Role of SAE Features}

In this experiment, the model performs enrichment without utilizing SAE features. Instead, a generic prompt is added:  \textit{“NOTE - think carefully before answering.”}. This approach isolates the contribution of SAE features to performance.

\paragraph{Rationale:} This test evaluates the significance of SAE features in enhancing the model's reasoning capabilities.



\subsection{Experiment C - Analysing the Impact of the Entropy Component}

In this experiment, the model performs only one loop for all questions, analysing instances that would not have received enrichment under the \framework pipeline.

\paragraph{Rationale:} This experiment targets scenarios where enrichment was skipped due to the entropy threshold, testing the hypothesis that enrichment in these cases might impair performance.

\section{Discussion}
\label{sec:discussion}

The results demonstrate the effectiveness of \framework in mitigating hallucinations and improving LLM performance. As shown in Table~\ref{tab:overall_results}, \framework consistently improves accuracy across three diverse datasets. Notably, Table~\ref{tab:larger_llm} indicates that enhancing a smaller model with \framework can yield performance comparable to or better than its larger counterpart. The only exception is the LLaMA model on BioASQ, where the 70B variant significantly outperforms the 8B model with \framework. However, \framework still provides measurable gains for the smaller model, underscoring its practical utility.

Ablation results in Table~\ref{tab:ablations} highlight the contributions of key components. Experiments A1 and A2 show that detecting and removing misleading features improves performance, though simply discarding them is insufficient when such features are absent. Conversely, focusing solely on reliable features can overly narrow the model's attention. Experiment B demonstrates that naive reflective prompting fails to match our enrichment strategy, except for minor improvements with Gemma2-9b on select datasets. Experiment C confirms the importance of entropy-based uncertainty estimation, as indiscriminate enrichment degrades performance.

Together, these findings demonstrate that \framework's synergy of entropy-driven detection and SAE-guided enrichment enhances LLM reliability without additional model training.


\section{Conclusion}

Hallucination remains a persistent challenge in LLM-based applications, undermining their reliability and trustworthiness in real-world deployments. In this work, we propose \frameworkv, a Sparse Autoencoder-based Framework for Robust Query Enrichment, which mitigates hallucinations by refining input queries and guiding model responses through interpretable, semantically grounded feature selection. \framework employs a two-stage process: first, it detects hallucinations using entropy-based uncertainty estimation; subsequently, it mitigates these issues by enriching queries with features derived from a SAE. Empirical evaluations across diverse benchmark datasets demonstrate that \framework significantly reduces hallucination rates while improving response accuracy by up to 29.45\%. Ablation studies confirm the critical role of entropy-based detection and SAE-driven enrichment in achieving these gains. Notably, \framework operates in a training-free manner, offering a lightweight, plug-and-play solution that seamlessly integrates into existing LLM pipelines without the need for additional model fine-tuning.


\label{sec:conclusion}

\section{Limitations}

While \framework demonstrates promising results in hallucination mitigation, it comes with certain limitations. First, the reliance on an SAE and the availability of auto-interpretable features constrains its applicability to LLMs that expose such internal representations. Extending the approach to models without these characteristics would require modifications or alternative interpretability techniques. Second, the effectiveness of the method is inherently influenced by the quality of the input queries. Although this is a common challenge across LLM-based systems, we explicitly acknowledge it here, as low-quality queries may still lead to suboptimal performance. Nevertheless, our evaluation on benchmark datasets, which span diverse query distributions, underscores the robustness and generalizability of our framework. Finally, our current implementation is restricted to English-language inputs, leaving multilingual and multimodal extensions as promising directions for future research.

\label{sec:limitations}

\clearpage

\bibliography{custom}



\end{document}